\documentclass[sigconf,nonacm]{acmart}

\AtBeginDocument{}
\setcopyright{none}
\copyrightyear{2026}
\acmYear{2026}

\usepackage{booktabs}
\usepackage{graphicx}
\usepackage{amsmath}   % NOT amssymb: acmart's newtxmath already defines its symbols (\Bbbk clash)
\usepackage{xcolor}
\usepackage{tikz}
\usetikzlibrary{arrows.meta,calc}
\definecolor{tgtc}{RGB}{31,84,150}   % target / prediction path in Figure 2
\definecolor{uotc}{RGB}{171,85,17}   % unbalanced-transport branch in Figure 2

\begin{document}

\title{OrderFlowDiT: Transport-Regularized Diffusion Transformer for Probabilistic Limit-Order Supply Forecasting}

%% arXiv is not anonymous: list every author. Add the advisor as a second \author block
%% (same \affiliation; the \email line is optional). Add \orcid{...} only with a real ORCID.
\author{Jung Hun Phee}
%\email{jphee3@gatech.edu}
\affiliation{%
  \institution{Georgia Institute of Technology}
  \city{Atlanta}
  \state{Georgia}
  \country{USA}}

\author{Tucker Balch}
%\email{교수님이메일@gatech.edu}   % 이메일을 안 넣기로 하면 이 줄만 지우세요
\affiliation{%
  \institution{Emory University}
  \city{Atlanta}
  \state{Georgia}
  \country{USA}}

\renewcommand{\shortauthors}{Phee and Balch}

\begin{abstract}
% Problem
We study the conditional distribution of limit-order volume submitted over
the next five minutes, given the current order book and recent order flow.
The task requires capturing price placement, total volume, and predictive
uncertainty jointly.
% Gap
Coordinate-wise diffusion losses do not explicitly account for price
displacement, while normalizing fields to unit mass removes information
about total volume. Agreement on selected summary statistics also does not
establish fidelity to the joint field distribution.
% Method and contributions
We introduce OrderFlowDiT, a conditional Diffusion Transformer for
unnormalized bid and ask volume-at-price fields, trained with an unbalanced
optimal-transport (UOT) auxiliary loss. We also develop a full-field
evaluation framework that combines a transport cost sensitive to price
placement and total volume with proper scoring rules and calibration
diagnostics.
% Experimental scope and results
In the primary comparison on $1{,}950$ validation conditions across $30$
U.S. equities, adding UOT reduces full-field transport cost by $45.8\%$ and
the transport energy score by $26.1\%$ relative to training with mean
squared error alone. Both measures improve for every stock.
% Implication
These findings support incorporating price geometry and volume into
generative modeling, with field accuracy and predictive uncertainty
evaluated together.
\end{abstract}


\maketitle

\section{Introduction}
\label{sec:introduction}

Liquidity in limit-order-book (LOB) markets is dynamic: trading over a
finite horizon depends on incoming liquidity as well as the orders
currently resting in the book. Models of optimal execution show that
liquidity replenishment rates affect trading schedules and execution
costs~\cite{obizhaeva2013optimal}. Empirical studies document
state-dependent limit-order submissions and the recovery of liquidity
following trades and
cancellations~\cite{biais1995empirical,lo2015resiliency}. Moreover,
estimated quote responses to incoming limit orders vary with their size
and price aggressiveness, highlighting the importance of both submitted
volume and price placement~\cite{hautsch2012market}. A forecast of the
book at the end of a horizon, however, does not separately identify the
new volume submitted during that interval, because submissions can be
offset by cancellations and executions. These observations motivate
forecasting how much new limit-order volume will arrive and where it
will be submitted, as one component of the future trading environment.

Forecasting future order supply requires characterizing uncertainty
in both submitted volume and its price allocation. Empirical
order-flow analysis documents interdependence between order sizes
and arrival dynamics~\cite{rambaldi2017role}. A single expected
submission profile cannot express this uncertainty: two predictive
distributions can have the same mean field while assigning different
probabilities to low submitted volume within a given price region.
This motivates representing forecasts as predictive distributions,
rather than point predictions alone~\cite{gneiting2007probabilistic}.
Moreover, marginal distributions at individual prices do not determine
how volumes vary jointly across prices and sides; constructing
multivariate forecast scenarios requires accounting for dependence
as well as the marginal
distributions~\cite{schefzik2013uncertainty}. Total-volume forecasts
omit price allocation, whereas unit-mass price profiles discard
absolute volume. We therefore seek a conditional joint distribution
over unnormalized bid and ask volume-at-price fields, allowing total
side volumes, price-localized submissions, and bid--ask submission
asymmetries to be examined within the same generated scenarios.

Specifically, we model the conditional joint distribution of bid and
ask limit-order submission fields over the next five minutes, given
the current LOB snapshot and recent order flow. Using both book state
and flow history is consistent with queue-reactive Hawkes models that
explicitly combine these sources of
information~\cite{wu2019queuereactive}. Each field coordinate records
the cumulative volume of new limit orders submitted on one side within
a price bin during the forecast window. We fix the price grid relative
to the mid-price observed at the forecast origin and retain absolute
submitted volume without unit-mass normalization. Unlike message-level
generators coupled to a book
simulator~\cite{nagy2023lobs5}, our target aggregates submissions without
retaining event order. Since order-book dynamics also involve
cancellations and executions~\cite{huang2015queuereactive}, the
submission fields alone do not determine the subsequent book state
or remaining liquidity. We therefore position this conditional
order-supply model as a potential component of a financial world
model, rather than as a complete simulator of order-book evolution.

Learning these submission fields calls for an objective sensitive to
both price displacement and differences in submitted volume. We use
a conditional diffusion model with a denoising mean squared error
(MSE) objective~\cite{ho2020denoising}. This objective penalizes errors
coordinatewise, without explicitly assigning costs to displacement
between price levels. Optimal transport provides a geometry-aware
comparison by matching mass across locations according to a ground
cost~\cite{feydy2019interpolating}. However, balanced transport requires
equal total mass, whereas predicted and observed submission fields
can have different total volumes. Unbalanced optimal transport (UOT)
relaxes the marginal constraints through divergence penalties,
allowing spatial displacement and mass discrepancies to be considered
jointly~\cite{sejourne2019sinkhorn}. We therefore augment diffusion
MSE with a debiased unbalanced Sinkhorn divergence on decoded
denoised fields, introducing an auxiliary training signal that
explicitly accounts for price placement and unequal submitted volume.

Evaluating these forecasts requires assessing not only discrepancies
between generated and observed fields, but also the quality of the
predictive distribution. Comparisons of market statistics, such as
those included in LOB-Bench~\cite{nagy2025lobbench}, provide useful
measures of realism, but agreement on selected summaries alone does
not determine the conditional joint distribution of submission
fields. Even when full fields are compared, averaging distances from
individual generated samples to the realization can favor overly
concentrated forecasts rather than faithful representations of
uncertainty~\cite{gneiting2007strictly}. Proper scoring rules instead
evaluate predictive distributions and are optimized in expectation
by the data-generating
distribution~\cite{gneiting2007strictly}. We therefore combine
mass-aware full-field transport comparisons with proper scoring
rules and calibration
diagnostics~\cite{gneiting2007probabilistic}, evaluating forecasts
against their corresponding realizations at matched forecast
origins and horizons.

To address this task, we introduce OrderFlowDiT, a conditional Diffusion
Transformer for the joint generation of unnormalized bid and ask
limit-order submission fields, together with a training objective and
evaluation framework tailored to price allocation, submitted volume,
and predictive uncertainty. Our contributions are threefold:
\begin{enumerate}
    \item We incorporate an unbalanced optimal-transport (UOT)
    auxiliary objective into conditional diffusion training.
    A debiased unbalanced Sinkhorn divergence on decoded denoised
    fields augments the diffusion mean squared error (MSE),
    providing a field-level training signal that explicitly accounts
    for price displacement and differences in total submitted volume.

    \item We construct a full-field evaluation framework that combines
    a mass-aware transport cost with transport and joint-field energy
    scores, the continuous ranked probability score (CRPS) for
    side volumes, and calibration diagnostics. These complementary
    measures assess field accuracy and predictive uncertainty at
    matched forecast origins and horizons.

    \item We compare MSE-only and MSE+UOT training with statistical
    baselines across $1{,}950$ forecast conditions from $30$ U.S.
    equities, reporting pooled and per-stock results. We examine
    accuracy--diversity tradeoffs and volume calibration in a
    Group Relative Policy Optimization (GRPO) fine-tuned variant,
    and assess a larger training configuration that jointly increases
    model capacity, training data, and optimization budget.
\end{enumerate}

Across 1,950 validation conditions on 30 U.S. equities, Base-UOT lowers full-field transport cost by $45.8\%$ and transport energy score by $26.1\%$ versus Base-MSE, with both improving for every stock. Joint-field energy and log-side-volume CRPS also improve. The evaluated GRPO configuration further improves transport energy and volume CRPS but worsens full-field cost, joint-field energy, and pooled volume-rank calibration while reducing ensemble dispersion, highlighting trade-offs across distributional metrics.
\section{Related Work}
\label{sec:related-work}

\paragraph{Market simulation and conditional order-flow generation.}
Market simulation relies on a model of the order flow that drives changes
in liquidity. Agent-based platforms such as ABIDES generate this flow
through interactions among trading agents and an exchange~\cite{byrd2020abides},
while queue-reactive models specify event intensities conditional on book
state~\cite{huang2015queuereactive}. Learning aggregate trading behavior
directly from historical data provides a complementary approach, as in the
conditional generators Stock-GAN and World
Agent~\cite{li2020stockgan,coletta2022worldagent}. Autoregressive models learn
the corresponding message sequences~\cite{nagy2023lobs5}, and MarS integrates
order and order-batch generation within an interactive
simulator~\cite{li2025mars}. At the level of book states, DiffVolume uses
conditional diffusion to generate future resting-volume
snapshots~\cite{wang2025diffvolume}. Our work addresses future liquidity
provision through the conditional generation of new limit-order volume
accumulated at each price over a five-minute window. Given the observed
book and recent flow, the model produces fields that preserve price
placement and total submitted volume while abstracting from within-window
event sequences.

\paragraph{Optimal transport for generative training.}
Optimal transport incorporates the geometry of the data into distributional
comparisons through a ground cost~\cite{peyre2019computational}. Entropic
regularization makes these comparisons computationally tractable, enabling
generative models to be trained with Sinkhorn-based
losses~\cite{genevay2018sinkhorn}. For sequential data, COT-GAN further imposes
causal constraints on transport couplings to respect temporal
structure~\cite{xu2020cotgan}. These approaches compare distributions of
generated and observed samples. Our training objective applies transport
within each prediction--target pair, treating submitted volumes as measures
supported on the price grid. Because total submitted volume varies across
fields, this comparison motivates an unbalanced formulation that accommodates
unequal mass~\cite{piccoli2014generalized}. Debiased unbalanced Sinkhorn
divergences provide differentiable losses for this
setting~\cite{sejourne2019sinkhorn}. We incorporate such a divergence as an
auxiliary objective alongside the diffusion denoising loss, making training
sensitive to both price displacement and volume mismatch.

\paragraph{Market realism and predictive-distribution evaluation.}
Market simulators are commonly evaluated by their ability to reproduce
stylized facts of returns and order-book
dynamics~\cite{cont2001stylized,vyetrenko2020getreal}. LOB-Bench formalizes
this approach through $L_1$ and Wasserstein-1 comparisons of conditional and
unconditional market-statistic distributions, complemented by a learned
discriminator and event-response diagnostics~\cite{nagy2025lobbench}.
Agreement on selected scalar summaries alone does not determine the joint
distribution of volume across prices. We complement these diagnostics by
comparing complete, unnormalized submission fields at matched forecast
boundaries, with a transport-plus-volume cost that retains price geometry
and total volume. A low average cost across generated draws, however, can
favor concentrated forecasts without correctly representing predictive
uncertainty. Proper scoring rules provide a principled basis for evaluating
the predictive distribution~\cite{gneiting2007strictly}, and fair ensemble
estimators account for finite sampling~\cite{ferro2014fair}. We therefore
pair the field cost with transport and joint-field energy scores,
supplemented by the continuous ranked probability score (CRPS) and rank
diagnostics for total volume. Section~\ref{sec:metric} specifies the support
and propriety conditions for each score.

\paragraph{Reward-based diffusion fine-tuning.}
Denoising diffusion policy optimization (DDPO) treats iterative denoising
as a sequential decision process, enabling policy-gradient updates
from rewards assigned to generated samples~\cite{black2024ddpo}.
Group Relative Policy Optimization (GRPO), introduced for mathematical
reasoning in language models, estimates advantages from rewards
normalized within groups of outputs sampled under the same
condition~\cite{shao2024deepseekmath}.
DanceGRPO applies this approach to diffusion and rectified-flow models
for image and video generation~\cite{xue2025dancegrpo}.
We apply group-relative fine-tuning to conditional order-flow generation,
using a reward that combines price displacement, volume error, and
diversity among generated fields.
Our evaluation examines whether improvements under the training reward
extend to distributional accuracy and volume calibration.

\section{Problem Formulation and Data Representation}
\label{sec:problem}

We study the conditional distribution of new limit-order volume submitted during the
next time window. The prediction target is a pair of volume-at-price fields, one for
each side of the book, conditioned on an observed book snapshot and recent order flow.

\subsection{Book state and submission fields}
The book state records resting liquidity at an instant. Let $\theta>0$ be
the tick size and $s\in\{+1,-1\}$ index bids and asks, respectively.
We write $X_t(p,s)$ for the number of shares resting at price
$p\in\theta\mathbb{Z}$ on side $s$ immediately before time $t$, reconstructed
from the preceding message history. Its best bid $b_t$ and best ask $a_t$
define the mid-price $m_t=(b_t+a_t)/2$.

The submission field records new limit-order volume accumulated over an
interval. For each new submission $i$, let $t_i$, $p_i$, $s_i$, and $q_i$
denote its arrival time, price, side, and volume in shares. For an interval
$W$, define
\begin{equation}
    \label{eq:field}
    A^{W}_s(p)=\sum_{i:\,t_i\in W}
        q_i\,\mathbf{1}\{s_i=s,\ p_i=p\},
\end{equation}
so $A^W=(A^W_{+1},A^W_{-1})$ is a pair of non-negative, unnormalized
measures on the price grid. The side mass
$M_s(A^W)=\sum_p A^W_s(p)$ is the total submitted volume on side $s$.
Retaining volumes in shares preserves both price placement and the amount
of liquidity submitted.

Each submission contributes its volume even if it is later canceled or
executed. Aggregation discards order identities and within-window event
order. Bid and ask supports may overlap: as quotes move, a price can
receive submissions on different sides at different times. Cancellation
and execution flows are outside the target, so a generated submission
field alone does not determine the next book state.

\subsection{Conditional forecasting task}
\begin{figure}[t]
    \centering
    \includegraphics[width=\columnwidth]{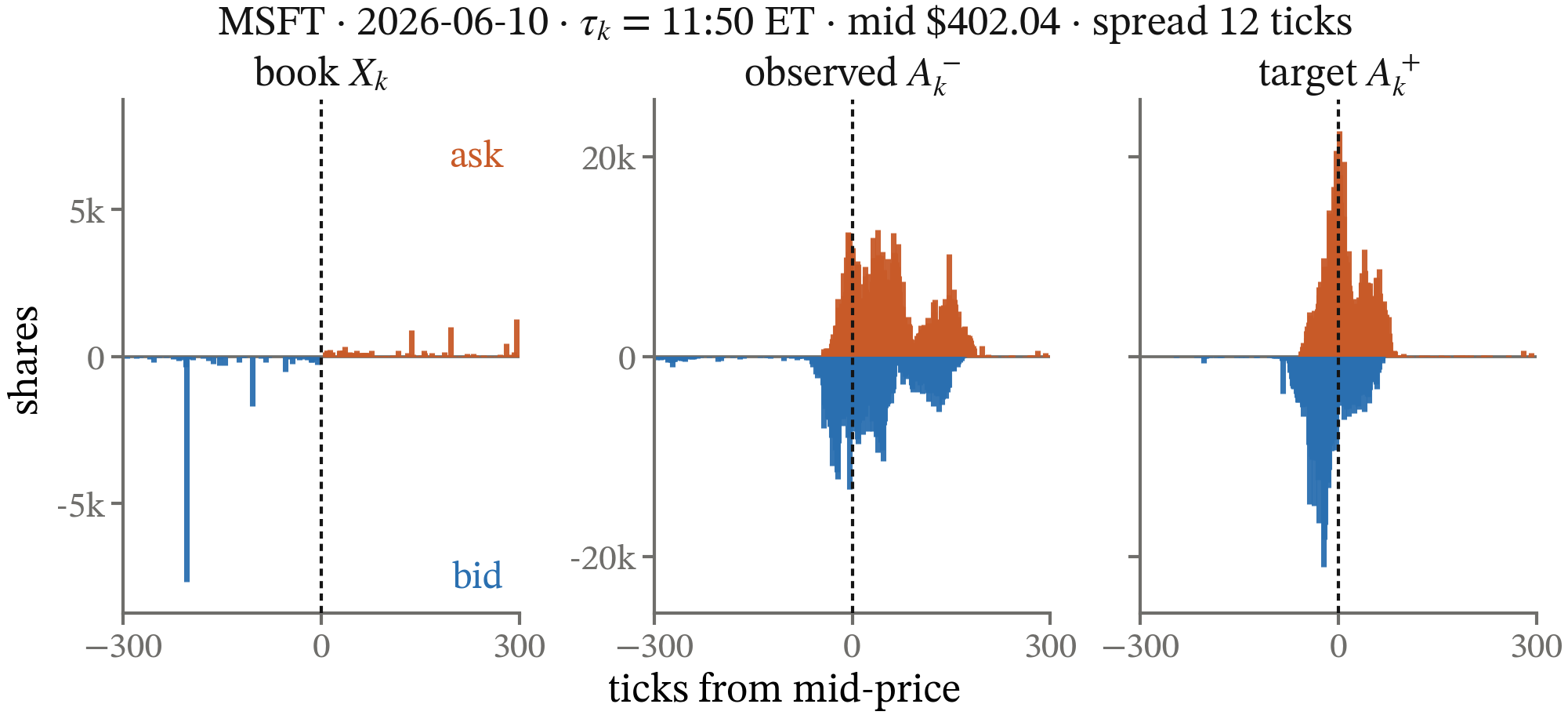}
    \caption{The data at one forecast boundary, within $300$ ticks of the
    mid-price: the book snapshot $X_k$, the observed field $A_k^-$, and the
    target field $A_k^+$. Ask volume is drawn upward and bid volume downward;
    both are non-negative, unnormalized share counts, and the two flow panels
    share a vertical scale. This view holds $53\%$ of the submitted volume and
    $1.2\%$ of the resting depth; the rest lies farther from the mid.}
    \Description{Three panels side by side: the book snapshot, the observed
    submission field, and the target submission field, each plotted against
    price offset from the mid-price over a range of three hundred ticks either
    side.}
    \label{fig:data-fields}
\end{figure}

At each forecast boundary $\tau_k$, the observed field summarizes
submissions in the preceding window, and the target covers the following
window. For a common duration $\Delta w$, these fields are
\begin{equation}
    \label{eq:flow-windows}
    A_k^-=A^{[\tau_k-\Delta w,\,\tau_k)},\qquad
    A_k^+=A^{[\tau_k,\,\tau_k+\Delta w)}.
\end{equation}
We use $\Delta w=300$ seconds (five minutes) in our experiments.
Figure~\ref{fig:data-fields} shows $X_k$, $A_k^-$ and $A_k^+$ at one boundary.

% Three interchangeable exhibits live in figures/, ordered by how far the target
% departs from the observed window (shape-only sided W_1, in ticks):
%   data_fields_typical_MSFT_20260623_1300.png    34 ticks, 84th pct   (view +/-100)
%   data_fields_moderate_MSFT_20260610_1150.png   48 ticks, 93rd pct   (view +/-300)  <- in use
%   data_fields_extreme_MSFT_20260609_1045.png   215 ticks,  1st of 7747 (view +/-300)
% Swapping one in means swapping its caption too; ready-to-paste captions and the
% supporting numbers are in
% studies/20260930_icaif_data_figure/outputs/figure_options.md

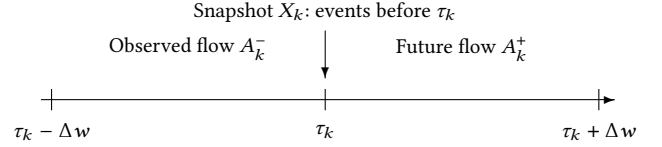
\begin{figure}[t]
    \centering
    \setlength{\unitlength}{1pt}
    \begin{picture}(234,65)
        \put(10,26){\vector(1,0){216}}
        \put(14,22){\line(0,1){8}}
        \put(117,22){\line(0,1){8}}
        \put(220,22){\line(0,1){8}}
        \put(14,40){\makebox(103,10){\small Observed flow $A_k^-$}}
        \put(117,40){\makebox(103,10){\small Future flow $A_k^+$}}
        \put(14,9){\makebox(0,10){\small $\tau_k-\Delta w$}}
        \put(117,9){\makebox(0,10){\small $\tau_k$}}
        \put(220,9){\makebox(0,10){\small $\tau_k+\Delta w$}}
        \put(117,54){\makebox(0,10){\small Snapshot $X_k$: events before $\tau_k$}}
        \put(117,52){\vector(0,-1){18}}
    \end{picture}
    \caption{Forecast timing in the contemporaneous-snapshot setting. All windows are
    left-closed and right-open; the target begins at the forecast boundary.}
    \Description{A timeline with two adjacent windows of equal duration. Observed
    submission flow occupies the window before the forecast boundary, and future
    submission flow occupies the window after it. The book snapshot at the boundary
    contains only events with earlier timestamps.}
    \label{fig:forecast-timeline}
\end{figure}

In addition to $A_k^-$, the conditioning information comprises the book
snapshot $X_k\equiv X_{\tau_k}$, summary features $\mathbf{h}_k$ for its
bid--ask spread and the time of session, and the instrument identity
$\mathrm{sym}_k$. Both $X_k$ and $A_k^-$ use messages strictly before
$\tau_k$. A new submission at exactly $\tau_k$ contributes only to the
target field $A_k^+$ (Figure~\ref{fig:forecast-timeline}).

The forecasting task is to learn the conditional distribution
\begin{equation}
    \label{eq:task}
    p\!\left(A_k^+\,\middle|\,X_k,A_k^-,\mathbf{h}_k,\mathrm{sym}_k,\Delta w\right).
\end{equation}
This distribution jointly describes submitted volume across prices and
both sides of the book. For each condition, a generative model produces
$R$ sampled fields representing possible future outcomes. We compare this
ensemble with the realized field $A_k^+$ using the evaluation measures in
Section~\ref{sec:metric}.

\subsection{Mid-relative representation and preprocessing}
\label{sec:model-rep}
We represent the book, observed flow, and future submission field on a
common price grid anchored to the observed snapshot. Let
$q_k=\lfloor m_{\tau_k}/\theta\rfloor$ be the snapshot mid-price rounded
down to an integer tick. The grid contains $B=2K+1$ bins, with
\begin{equation}
    \label{eq:price-grid}
    p_{k,b}=\theta(q_k+b-K),\qquad b=0,\ldots,B-1.
\end{equation}
Thus each bin refers to the same price across all fields in a prediction
example. We use $K=5000$, giving $B=10001$ bins at one-tick spacing.
The model predicts within this finite grid. Volume outside the grid is
recorded for coverage diagnostics, and evaluation retains the full
support of the realized target.

For numerical conditioning, we apply a logarithmic transform followed
by standardization:
\begin{equation}
    \label{eq:transform}
    T(v)=\frac{\log(1+v)-\mu}{\tilde{\sigma}},
    \qquad \tilde{\sigma}=\max\{\sigma,10^{-6}\}.
\end{equation}
The mean $\mu$ and standard deviation $\sigma$ are estimated separately
for bid and ask book depth and submission flow, using training cells
including zeros. Generated values are decoded through the inverse
transform and clipped to non-negative share volumes before evaluation.
No unit-mass normalization is applied, so total submitted volume remains
part of the prediction.

On this grid, the transformed target
$Y=T(A_k^+)\in\mathbb{R}^{2\times B}$ contains bid and ask submission flow.
The conditioning tensor $C_k\in\mathbb{R}^{6\times B}$ combines
transformed bid and ask book depth, two indicator channels marking the
best quotes, and transformed bid and ask observed flow.
The summary vector $\mathbf{h}_k$ contains the spread in ticks and
sine--cosine features for the time of session, standardized using
training statistics. Instrument identity and the window duration
provide additional global conditioning inputs.

Data coverage, chronological splits, and the evaluation protocol are
given in Section~\ref{sec:experiments}.

\section{Conditional Order-Flow Model}
\label{sec:model}

\begin{figure*}[t]
\centering
\begin{tikzpicture}[
  x=1cm, y=1cm, font=\footnotesize,
  >={Stealth[length=4pt,width=3.2pt]},
  blk/.style={draw=black!55, rounded corners=1.2pt, fill=black!4,
              align=center, inner xsep=3pt, inner ysep=2.5pt, line width=0.4pt},
  tgtblk/.style={blk, fill=tgtc!12, draw=tgtc!75},
  uotblk/.style={blk, fill=uotc!13, draw=uotc!80},
  strip/.style={draw=black!45, line width=0.35pt},
  bar/.style={black!50, line width=0.42pt},
  tbar/.style={tgtc!70, line width=0.42pt},
  tok/.style={draw=black!55, fill=black!7, minimum width=0.20cm,
              minimum height=0.28cm, inner sep=0, line width=0.35pt},
  fl/.style={->, black!65, line width=0.5pt},
  dfl/.style={->, black!50, line width=0.4pt, dotted},
  mod/.style={->, black!60, line width=0.5pt, dash pattern=on 2.2pt off 1.2pt on 0.5pt off 1.2pt},
  loop/.style={->, tgtc!75, line width=0.55pt, dashed},
  rl/.style={font=\scriptsize, inner sep=1pt, anchor=east},
  sl/.style={font=\scriptsize, inner sep=1pt, align=center},
  tag/.style={font=\scriptsize\itshape, inner sep=1pt, text=black!60},
]

%% ================= Panel A: fields aligned on one price grid =================
\def\sx{1.20}\def\sw{3.55}\def\mid{2.975}

\foreach \yy/\nm/\sd in {4.62/{bid $Z_n$}/17, 4.32/{ask $Z_n$}/43} {
  \node[rl] at (\sx-0.08,\yy+0.14) {\nm};
  \begin{scope}[shift={(\sx,\yy)}]
    \fill[tgtc!5] (0,0) rectangle (\sw,0.28);
    \draw[strip] (0,0) rectangle (\sw,0.28);
    \pgfmathsetseed{\sd}
    \foreach \i in {0,...,57} {
      \pgfmathsetmacro{\dx}{0.035+0.0605*\i}
      \pgfmathsetmacro{\h}{0.035+0.215*rnd}
      \draw[tbar] (\dx,0.012) -- (\dx,{0.012+\h});}
  \end{scope}}

\node[rl] at (\sx-0.08,4.06) {bid $X_k$};
\begin{scope}[shift={(\sx,3.92)}]
  \fill[black!2] (0,0) rectangle (\sw,0.28);\draw[strip] (0,0) rectangle (\sw,0.28);
  \foreach \i in {0,...,57} {
    \pgfmathsetmacro{\dx}{0.035+0.0605*\i}
    \pgfmathsetmacro{\h}{\dx<1.72 ? 0.245*exp(-((\dx-1.64)/0.46)^2) : 0}
    \draw[bar] (\dx,0.012) -- (\dx,{0.012+\h});}
\end{scope}
\node[rl] at (\sx-0.08,3.76) {ask $X_k$};
\begin{scope}[shift={(\sx,3.62)}]
  \fill[black!2] (0,0) rectangle (\sw,0.28);\draw[strip] (0,0) rectangle (\sw,0.28);
  \foreach \i in {0,...,57} {
    \pgfmathsetmacro{\dx}{0.035+0.0605*\i}
    \pgfmathsetmacro{\h}{\dx>1.81 ? 0.245*exp(-((\dx-1.90)/0.46)^2) : 0}
    \draw[bar] (\dx,0.012) -- (\dx,{0.012+\h});}
\end{scope}
\node[rl] at (\sx-0.08,3.46) {bid quote};
\begin{scope}[shift={(\sx,3.32)}]
  \fill[black!2] (0,0) rectangle (\sw,0.28);\draw[strip] (0,0) rectangle (\sw,0.28);
  \draw[bar,line width=1.1pt] (1.685,0.012) -- (1.685,0.25);
\end{scope}
\node[rl] at (\sx-0.08,3.16) {ask quote};
\begin{scope}[shift={(\sx,3.02)}]
  \fill[black!2] (0,0) rectangle (\sw,0.28);\draw[strip] (0,0) rectangle (\sw,0.28);
  \draw[bar,line width=1.1pt] (1.865,0.012) -- (1.865,0.25);
\end{scope}
\node[rl] at (\sx-0.08,2.86) {bid $A_k^-$};
\begin{scope}[shift={(\sx,2.72)}]
  \fill[black!2] (0,0) rectangle (\sw,0.28);\draw[strip] (0,0) rectangle (\sw,0.28);
  \foreach \i in {0,...,57} {
    \pgfmathsetmacro{\dx}{0.035+0.0605*\i}
    \pgfmathsetmacro{\h}{\dx<1.74 ? 0.20*exp(-((\dx-1.52)/0.26)^2)+0.085*exp(-((\dx-0.72)/0.11)^2) : 0}
    \draw[bar] (\dx,0.012) -- (\dx,{0.012+\h});}
\end{scope}
\node[rl] at (\sx-0.08,2.56) {ask $A_k^-$};
\begin{scope}[shift={(\sx,2.42)}]
  \fill[black!2] (0,0) rectangle (\sw,0.28);\draw[strip] (0,0) rectangle (\sw,0.28);
  \foreach \i in {0,...,57} {
    \pgfmathsetmacro{\dx}{0.035+0.0605*\i}
    \pgfmathsetmacro{\h}{\dx>1.79 ? 0.20*exp(-((\dx-2.02)/0.26)^2)+0.085*exp(-((\dx-2.82)/0.11)^2) : 0}
    \draw[bar] (\dx,0.012) -- (\dx,{0.012+\h});}
\end{scope}

\draw[tgtc!65, dashed, line width=0.45pt] (\mid,2.36) -- (\mid,4.99);

\draw[->,black!70,line width=0.45pt] (\sx,2.18) -- (\sx+\sw+0.18,2.18);
\draw[black!70,line width=0.45pt] (\mid,2.10) -- (\mid,2.26);
\node[sl] at (\mid,1.93) {$q_k$ = snapshot mid bin\,; \ $B=10001$ bins, one tick wide};
\node[sl] at (\mid,1.66) {$Z_n$ ($2\times B$) and $C_k$ ($6\times B$) share this mid-anchored grid};

\draw[tgtc,line width=0.5pt] (2.10,2.36) rectangle (2.32,4.99);
\node[sl,text=tgtc] at (\mid,5.24) {patch of $P{=}16$ bins};
\draw[tgtc,line width=0.4pt] (2.50,5.14) -- (2.28,5.01);

%% ================= Panel B: tokens and transformer =================
\node[tgtblk] (etgt) at (5.85,4.62) {$E_{\mathrm{tgt}}$};
\node[blk]    (econd) at (5.85,3.05) {$E_{\mathrm{cond}}$};
\draw[dfl] (2.36,4.68) -- (etgt.west);
\draw[dfl] (2.36,3.10) -- (econd.west);

\node[circle,draw=black!60,inner sep=0.8pt,line width=0.4pt] (plus) at (6.95,3.84) {$+$};
\draw[fl] (etgt.east) -- (6.60,4.62) -- (6.60,3.98) -- (plus.north);
\draw[fl] (econd.east) -- (6.60,3.05) -- (6.60,3.70) -- (plus.south);
\node[sl,anchor=west] at (7.06,4.16) {$+\,\mathrm{PE}$};
\node[sl,text=black!65] at (5.85,3.84) {each patch};

\foreach \i in {0,...,4} {\node[tok] at ({7.60+0.28*\i},3.84) {};}
\node[sl] at (9.10,3.84) {$\cdots$};
\node[tok] at (9.45,3.84) {};
\draw[fl] (plus.east) -- (7.44,3.84);
\node[sl] at (8.40,3.34) {$L=626$ tokens, $d=512$};

\node[blk,minimum width=2.10cm,minimum height=1.05cm] (dit) at (10.85,3.84)
  {DiT block, $\times\,12$\\[-1pt]{\scriptsize self-attention $+$ MLP}\\[-1pt]{\scriptsize over the $L$ price tokens}};
\draw[fl] (9.62,3.84) -- (dit.west);

\node[blk,align=center] (head) at (12.85,3.84) {linear head\\[-2pt]{\scriptsize unpatchify}};
\draw[fl] (dit.east) -- (head.west);

\foreach \yy/\sd in {3.90/5, 3.56/29} {
  \begin{scope}[shift={(13.85,\yy)}]
    \fill[tgtc!7] (0,0) rectangle (1.05,0.26);\draw[strip] (0,0) rectangle (1.05,0.26);
    \pgfmathsetseed{\sd}
    \foreach \i in {0,...,16} {
      \pgfmathsetmacro{\dx}{0.03+0.0605*\i}
      \pgfmathsetmacro{\h}{0.03+0.20*rnd}
      \draw[tbar] (\dx,0.011) -- (\dx,{0.011+\h});}
  \end{scope}}
\draw[fl] (head.east) -- (13.85,3.84);
\node[sl,text=tgtc!85] at (14.37,3.26) {velocity $v_\varphi$};

%% ================= global conditioning =================
\node[blk,align=center,minimum width=4.5cm] (glob) at (9.30,0.78)
  {global conditioning vector\\[-1pt]{\scriptsize step $n$ $\cdot$ $\mathbf{h}_k$ (spread, time of session) $\cdot$ $\mathrm{sym}_k$ $\cdot$ $\log\Delta w$ $\cdot$ mean cond.\ token}};
\draw[mod] (10.85,1.22) -- (dit.south);
\node[sl,anchor=west] at (11.05,2.28) {AdaLN-Zero};
\node[sl,anchor=west] at (11.05,2.04) {shifts, scales, gates};

%% ================= objectives and sampling =================
\node[tag] at (16.45,5.28) {training objectives};
\node[blk,align=center,minimum width=2.35cm] (mse) at (16.45,4.80)
  {$\mathcal{L}_{\mathrm{MSE}}$ against $v$\\[-1pt]{\scriptsize masked valid bins}};
\draw[fl] (14.90,4.03) -- (15.10,4.03) -- (15.10,4.80) -- (mse.west);

\node[blk,align=center,minimum width=2.35cm] (yhat) at (16.45,2.85)
  {{\scriptsize$\hat Y=\alpha_nZ_n-\sigma_nv_\varphi$}\\[-1pt]{\scriptsize$\hat A=T^{-1}(\hat Y)$ in shares}};
\draw[fl] (14.90,3.69) -- (15.10,3.69) -- (15.10,2.85) -- (yhat.west);

\node[uotblk,align=center,minimum width=2.35cm] (uot) at (16.45,1.45)
  {{\scriptsize$\lambda_{\mathrm{UOT}}\alpha_n^2\,\bar S(\hat A_s,A^+_{k,s})$}\\[-1pt]{\scriptsize unbalanced transport}};
\draw[fl] (yhat.south) -- (uot.north);
\draw[fl] (13.55,1.45) -- (uot.west);
\node[sl,anchor=east] at (13.50,1.45) {realized $A_k^+$};
\node[tag,anchor=north] at (16.45,1.07) {Base-UOT only};

\draw[loop] (14.37,4.20) -- (14.22,5.62) -- (4.50,5.62) -- (4.50,4.94);
\node[sl,text=tgtc!85,anchor=south] at (9.40,5.70)
  {\itshape sampling: $50$ DDIM steps ($\eta=0$); independent initial noise per draw gives the $R$-member ensemble};

\end{tikzpicture}
\caption{OrderFlowDiT-Base. Every field shares the one-tick, mid-anchored
price grid of Section~\ref{sec:model-rep}, so a bin refers to the same price in
the noisy target $Z_n$ and in all six conditioning channels of $C_k$: book
depth $X_k$, the two best-quote indicators, and observed flow $A_k^-$. Patches
of $P=16$ bins are embedded separately for target and conditioning and summed;
bins padding the grid to whole patches are masked. Base-MSE trains on the
masked velocity loss alone, whereas the Base-UOT term is evaluated on the
decoded field $\hat A$ in shares rather than in velocity space. GRPO
(Section~\ref{sec:model-grpo}) fine-tunes this same network.}
\Description{A left-to-right schematic of the conditional diffusion
transformer. On the left, eight horizontal strips stacked over a shared
mid-anchored price axis show the two noisy target channels above the six
conditioning channels: bid and ask book depth, bid and ask best-quote
indicators, and bid and ask observed flow. A highlighted vertical slice marks
one patch of sixteen price bins. Separate patch embeddings for target and
conditioning are summed with a positional encoding to form a sequence of price
tokens, which pass through twelve transformer blocks and a linear head that
outputs a two-channel velocity field. A global conditioning vector, built from
the diffusion step, summary features, instrument identity, window duration and
the mean condition token, modulates each block through AdaLN-Zero. On the
right, the velocity feeds the masked mean-squared-error objective, and a second
branch decodes it to a predicted field in shares that is compared with the
realized field by an unbalanced transport term used only by the UOT variant. A
dashed arrow returning to the target channels marks the fifty-step DDIM
sampling loop.}
\label{fig:orderflowdit}
\end{figure*}
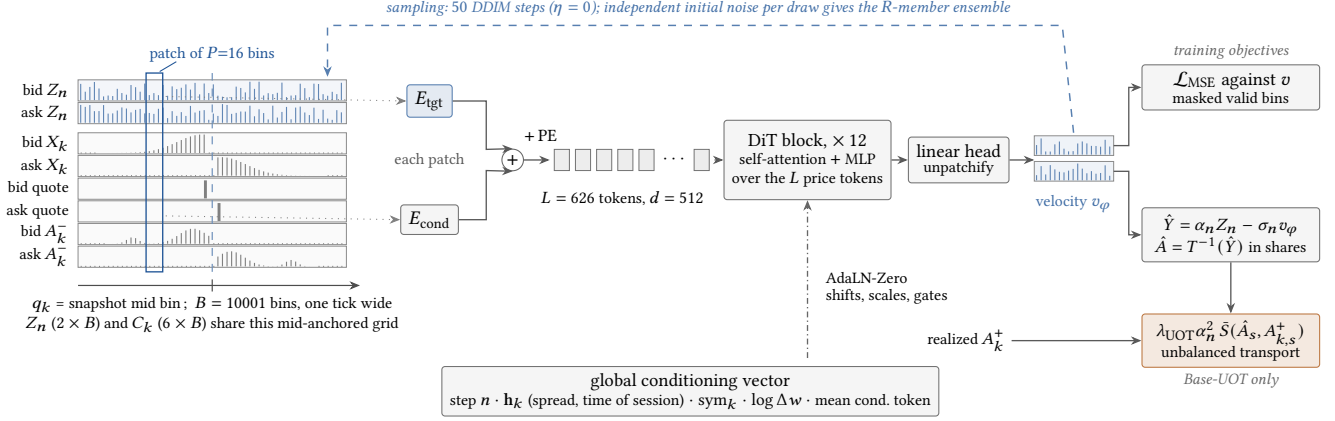

OrderFlowDiT generates the next-window submission field conditioned on the
current book and recent flow, using the representation in
Section~\ref{sec:model-rep}. Its MSE and UOT variants share a conditional
Diffusion Transformer; the UOT variant adds a transport objective on decoded
fields. We first define the diffusion model and its MSE loss, then introduce
the transport objective, sampling, and additional GRPO fine-tuning.
Figure~\ref{fig:orderflowdit} illustrates the Base configuration used in the
primary comparison.

\subsection{Conditional Diffusion Model}
\label{sec:model-net}
With $Y=T(A_k^+)$ and fixed conditioning
$c_k=(C_k,\mathbf{h}_k,\mathrm{sym}_k,\Delta w)$, training samples $n$
uniformly from a $1000$-level cosine schedule and independent noise
$\varepsilon\sim\mathcal{N}(0,I)$ to form
\begin{equation}
    \label{eq:forward}
    Z_n=\alpha_nY+\sigma_n\varepsilon,
    \qquad v=\alpha_n\varepsilon-\sigma_nY,
\end{equation}
where $\alpha_n^2+\sigma_n^2=1$.

Our price-axis DiT~\cite{peebles2023dit} sums separate linear embeddings of
aligned, non-overlapping $P$-bin patches from $Z_n$ and $C_k$ with fixed
sinusoidal positional encodings (Figure~\ref{fig:orderflowdit}). Bid and ask
enter jointly as channels; padding is masked. AdaLN-Zero uses summed
embeddings of $n$, $\mathbf{h}_k$, $\mathrm{sym}_k$, $\log\Delta w$, and the
mean valid conditioning token. A linear head predicts the velocity field
$v_\varphi(Z_n,n,c_k)$.

MSE training minimizes the expected masked velocity loss
\begin{equation}
    \label{eq:mse}
    \mathcal{L}_{\mathrm{MSE}}
    =\frac{\sum_{s,b}m_b\,[v_\varphi(Z_n,n,c_k)-v]_{s,b}^{\,2}}
           {2\sum_bm_b},
\end{equation}
where $m_b$ indicates valid price bins. Architecture and training settings
are given in Section~\ref{sec:exp-models}.

\subsection{Training with an Unbalanced Transport Objective}
\label{sec:model-diff}
The velocity MSE compares predictions coordinatewise. To account explicitly
for price displacement and unequal submitted volumes, we add an unbalanced
optimal-transport (UOT) objective on decoded fields. For each noisy example,
the implied clean prediction is
$\hat Y=\alpha_nZ_n-\sigma_nv_\varphi(Z_n,n,c_k)$.
We decode it to $\hat A=T^{-1}(\hat Y)$ in shares using the decoder in
Section~\ref{sec:model-sampling} and compare each side with the realized
in-grid field.

For non-negative volume vectors $u,v$, define entropic UOT with relaxed
marginals:
\begin{equation}
    \label{eq:uot}
    \begin{aligned}
    \mathcal{U}_{\epsilon,\rho}(u,v)
      =\min_{\pi\ge0}\ \big\{&
        \langle C,\pi\rangle+\epsilon\,\mathrm{KL}(\pi\|u\otimes v)\\
        &+\rho\,\mathrm{KL}(\pi\mathbf{1}\|u)
         +\rho\,\mathrm{KL}(\pi^\top\mathbf{1}\|v)\big\}.
    \end{aligned}
\end{equation}
The ground cost $C_{ij}=|i-j|$ measures displacement in ticks. The marginal
penalties allow the transport plan to accommodate unequal side volumes;
$\mathrm{KL}$ denotes generalized KL divergence, including its linear mass
terms. We use entropic regularization $\epsilon=4$ and marginal penalty
$\rho=20$. Following S\'ejourn\'e et al.~\cite{sejourne2019sinkhorn}, we remove
entropic self-bias:
\begin{equation}
    \label{eq:sinkhorn}
    \begin{aligned}
    S(u,v)={}&\mathcal{U}(u,v)
        -\tfrac12\mathcal{U}(u,u)-\tfrac12\mathcal{U}(v,v)\\
        &+\tfrac{\epsilon}{2}\,[M(u)-M(v)]^2,
    \end{aligned}
\end{equation}
where $M(u)=\sum_bu_b$ and $\mathcal{U}$ abbreviates
$\mathcal{U}_{\epsilon,\rho}$. To reduce variation in loss scale across
windows with different volumes, we divide by the pair's mean mass:
$\bar S(u,v)=S(u,v)/\max\{[M(u)+M(v)]/2,1\}$.
This scales the scalar loss while retaining both input fields in shares.
The combined training objective is
\begin{equation}
    \label{eq:total}
    \mathcal{L}
      =\mathbb{E}\!\left[
        \mathcal{L}_{\mathrm{MSE}}
        +\frac{\lambda_{\mathrm{UOT}}\alpha_n^2}{2}
          \sum_s\bar S(\hat A_s,A^+_{k,s})\right],
\end{equation}
with expectation over training examples, noise levels, and noise
realizations. The UOT variant uses $\lambda_{\mathrm{UOT}}=0.1$; setting it
to zero recovers MSE-only training. The factor $\alpha_n^2$ reduces the
auxiliary contribution at high noise levels, where clean-field estimates
are less reliable.

On the uniform grid, Sinkhorn kernel products use FFT convolution, with
detached dual potentials obtained from log-domain iterations. Evaluation
uses the distinct field distances and scoring rules in
Section~\ref{sec:metric}. The Base variants share a training split and
optimizer schedule; budgets and variant-specific checkpoint-selection
rules are given in Section~\ref{sec:exp-models}.

\subsection{Sampling and Predictive Ensembles}
\label{sec:model-sampling}
MSE and UOT ensembles use $R$ independent Gaussian initializations, each
sampled with $50$ DDIM steps~\cite{song2021ddim} ($\eta=0$), fixed
conditioning, and exponential moving average (EMA) weights. Final values
$y$ decode to $T^{-1}(y)=\max\{\exp(z)-1,0\}$ with
$z=\min\{\tilde{\sigma}y+\mu,32\}$. Fields retain non-negative, real-valued
share volumes without unit-mass normalization and map back to absolute
price ticks; non-positive-price bins are excluded and their mass recorded.

\subsection{GRPO Fine-Tuning}
\label{sec:model-grpo}
We further adapt Base-UOT with GRPO~\cite{shao2024deepseekmath}, optimizing
a reward on groups of generated fields while retaining the architecture
and representation. This \emph{Base-GRPO} variant treats stochastic DDIM
with $20$ steps and $\eta=1$ as a policy, drawing $G=8$ decoded fields $U_g$
per condition. For the realized field $A_k^+$, define the target-dependent
cost
\begin{equation}
    \label{eq:ell}
    \ell_k(U,V)=\frac12\sum_s\left[
       \frac{w_s(U,V)}{\kappa}
       +\beta_{\mathrm{mass}}\frac{|M_s(U)-M_s(V)|}{M_s(A_k^+)}
       \right],
\end{equation}
where $w_s$ is normalized-shape $W_1$ in ticks
(Section~\ref{sec:metric-point}), $\kappa=50$, and $\beta_{\mathrm{mass}}=1$.
For the reward, shape cost is zero when both sides are empty and $B/2$ ticks
when exactly one is empty; the mass term is omitted when $M_s(A_k^+)=0$.
Each draw receives reward
\begin{equation}
    \label{eq:reward}
    r_g=-\ell_k(U_g,A_k^+)
       +\frac{1}{2(G-1)}\sum_{g'\ne g}\ell_k(U_g,U_{g'}).
\end{equation}
The first term rewards agreement with the realization; the second
encourages separation between draws. Group-standardized rewards define
the advantages. Updates use a clipped policy-ratio surrogate and a KL
penalty to the frozen Base-UOT model, with log-ratios and KL terms
normalized by the number of valid scalar entries.

\section{Full-Field and Distributional Evaluation}
\label{sec:metric}

We evaluate both the accuracy of individual submission fields and the quality
of their predictive distribution. At forecast boundary $\tau_k$, independent
model draws $\widehat A_k^{(1)},\ldots,\widehat A_k^{(R)}$ are compared with
the realization $A_k^+$. These paired field comparisons complement
summary-statistic benchmarks such as LOB-Bench~\cite{nagy2025lobbench}.
All comparisons retain share volumes and the full supplied price support,
including realized flow outside the model's grid. A common price translation
leaves the scores unchanged, so absolute ticks and the shared mid-relative
frame give identical values. Table~\ref{tab:metrics} summarizes the eight
quantities; sampling and reporting follow Section~\ref{sec:exp-protocol}.

\subsection{Full-Field Accuracy}
\label{sec:metric-ffot}
\label{sec:metric-point}
For generic bid--ask fields $U,V$, write $j=p/\theta\in\mathbb{Z}$ for the
price tick and $M_s(U)=\sum_jU_s(j)$ for side mass. When this mass is positive,
let $F_{U,s}$ be the CDF of $U_s/M_s(U)$. The one-dimensional Wasserstein
shape discrepancy is
\begin{equation}
    \label{eq:metric-w1}
    w_s(U,V)=\sum_{j\in\mathbb{Z}}
        \big|F_{U,s}(j)-F_{V,s}(j)\big|.
\end{equation}
This exact CDF expression measures price displacement in ticks without a
transport solver~\cite{santambrogio2015ot}. Normalization removes total
volume, so fields with the same shape have zero discrepancy even when their
masses differ.

To retain both placement and volume, set $m_s=M_s(U)$ and $n_s=M_s(V)$ and
define
\begin{align}
    \label{eq:metric-dfull}
    d_{\mathrm{full},s}(U,V)
        &=\min(m_s,n_s)\,w_s(U,V)+\lambda|m_s-n_s|,\\
    \label{eq:metric-dfull-sides}
    d_{\mathrm{full}}(U,V)&=\sum_s d_{\mathrm{full},s}(U,V).
\end{align}
The first term transports the common mass between the normalized shapes;
the second charges for residual volume. With $\lambda=50$ ticks, the cost
has units of share$\cdot$ticks. An equal-mass displacement by $\delta$ ticks
costs $m_s|\delta|$; a volume change at fixed shape incurs the mass penalty.
If either mass is zero, the transport term is defined as zero, leaving
$d_{\mathrm{full},s}=\lambda|m_s-n_s|$.

This feasible plan leaves total marginal defect $|m_s-n_s|$, so its cost
upper-bounds unbalanced OT with a $\lambda$-weighted $L^1$ penalty on each
marginal discrepancy~\cite{piccoli2014generalized}. It need not attain the
optimal transport value or satisfy the triangle inequality. It also differs
from the entropic, KL-relaxed training objective in~\eqref{eq:uot}.

\paragraph{Error components.}
Alongside the combined cost, we report sided $W_1$, target-normalized
volume error $\mathrm{NTVE}_s(U,V)=|m_s-n_s|/n_s$ for a generated $U$ and
realization $V$, and log-field error
$\mathrm{MSE}_{\log}(U,V)=\rho_2(U,V)^2$, with $\rho_2$ defined
in~\eqref{eq:metric-rho-l2}. Sided $W_1$ and NTVE are zero when a side is
empty in both fields; sided $W_1$ is undefined when exactly one is empty
and NTVE when the target alone is empty, whereas $d_{\mathrm{full}}$ and
$\mathrm{MSE}_{\log}$ remain defined for empty fields.

\subsection{Predictive Distribution Quality}
\label{sec:metric-proper}
Averaging field costs over draws measures sample fidelity but does not
identify the predictive law: concentrated forecasts can achieve small
average errors. Proper scoring rules instead attain their minimum expected
value at the true conditional law; strict propriety makes that minimizer
unique~\cite{gneiting2007strictly}.

For a fixed condition, write $U^{(r)}=\widehat A_k^{(r)}$ and $V=A_k^+$.
Given a ground distance $\rho$ and $R\ge2$ independent draws, the fair
energy-score estimator is~\cite{ferro2014fair}
\begin{align}
    \label{eq:metric-es-hat}
    \widehat{\mathrm{ES}}_{\rho}
        &=\frac1R\sum_{r=1}^{R}\rho(U^{(r)},V)\notag\\
        &\quad-\frac{1}{2R(R-1)}
            \sum_{r\ne r'}\rho(U^{(r)},U^{(r')}).
\end{align}
The pairwise sum uses ordered pairs; independence makes the estimator
unbiased for the population score. We retain the mean target distance
$T_\rho$ and mean distance between distinct draws $D_\rho$, so
$\widehat{\mathrm{ES}}_\rho=T_\rho-D_\rho/2$.

\paragraph{Transport energy score.}
To combine price geometry with volume uncertainty, set
$g_s(U)=\log(1+M_s(U))$ and use
\begin{equation}
    \label{eq:metric-rho}
    \rho_T(U,V)=\frac12\sum_s
        \left[\frac{w_s(U,V)}{\kappa}
        +\beta\big|g_s(U)-g_s(V)\big|\right],
\end{equation}
with $\kappa=50$ ticks and $\beta=1$. The symmetric log-mass term reduces
the dominance of large volumes without exact scale invariance. On laws
with nonempty sides and finite first metric moments, $\mathrm{ES}_T$ is
proper because $\rho_T$ is a sum of $L^1$ distances and hence of negative
type~\cite{gneiting2007strictly,szekely2013energy}. Its additive side, shape,
and mass terms do not identify their dependence, so it is not strictly
proper for the joint field law. The GRPO reward in~\eqref{eq:reward} uses
a different mass discrepancy and does not inherit this propriety claim.

For each side, the target and all draws must be either all nonempty or all
empty; the latter receives zero score. Both side scores must be defined.
Mixed empty/nonempty ensembles are excluded, and the propriety statement
does not extend to laws mixing empty and nonempty fields.

\paragraph{Joint-field energy score.}
To assess the joint bid--ask law, define $z_s(U)(j)=\log(1+U_s(j))$ and
\begin{equation}
    \label{eq:metric-rho-l2}
    \rho_2(U,V)=
        \left[\frac{1}{2B}\sum_s\sum_{j\in\mathbb{Z}}
        \big(z_s(U)(j)-z_s(V)(j)\big)^2\right]^{1/2}.
\end{equation}
Here $B=10001$ fixes normalization without truncating support, and the log
transform uses raw shares without model-specific standardization.
The injective log-field map lies in a separable Hilbert space, whose norm
distance has strong negative type~\cite{lyons2013distance}. Thus the resulting
$\mathrm{ES}_2$ is strictly proper for the joint field law under finite
first metric moments, including laws with empty fields. Its binwise
geometry complements transport: the penalty for moving an isolated spike
to another bin does not increase with the distance moved.

\paragraph{Volume CRPS.}
Applying~\eqref{eq:metric-es-hat} to $|g_s(U)-g_s(V)|$ gives the fair
continuous ranked probability score (CRPS) for log side
volume~\cite{matheson1976scoring}; we average the two sides. Each score is
strictly proper for its log-volume marginal under a finite first
moment, including empty sides with $g_s=0$. Their average does not identify
bid--ask dependence.

\begin{table}[t]
    \caption{Evaluation quantities and their roles. Costs, errors, scores, and
    RI are lower-is-better. Propriety refers to population scores on the
    domains stated in the text.}
    \label{tab:metrics}
    \centering
    \small
    \setlength{\tabcolsep}{4pt}
    \begin{tabular}{@{}p{0.32\linewidth}p{0.62\linewidth}@{}}
        \toprule
        Quantity & Role \\
        \midrule
        $d_{\mathrm{full}}$ & Price placement and volume cost \\
        \midrule
        $\mathrm{ES}_T$ & Proper transport-based score \\
        $\mathrm{ES}_2$ & Strictly proper joint-field score \\
        Volume CRPS & Strictly proper side-volume score \\
        \midrule
        Sided $W_1$ & Shape diagnostic in ticks \\
        NTVE & Relative-volume diagnostic \\
        $\mathrm{MSE}_{\log}$ & Per-bin log-volume diagnostic \\
        \bottomrule
    \end{tabular}
\end{table}

\section{Experimental Setup}
\label{sec:experiments}

\subsection{Data and Experimental Splits}
\label{sec:exp-data}
We use LOBSTER full-depth order-message data for a fixed panel of $30$ U.S.
equities.\footnote{AAPL, AMGN, AMZN, AXP, BA, CAT, CRM, CSCO, CVX, DIS, GS,
HD, HON, IBM, JNJ, JPM, KO, MCD, MMM, MRK, MSFT, NKE, NVDA, PG, SHW, TRV,
UNH, V, VZ, and WMT.}
Each session is restricted to 10:30--16:00 Eastern Time, with a common
one-cent price lattice. We construct the contemporaneous-snapshot
(lag0) conditions of Section~\ref{sec:problem}: the book snapshot and
preceding five minutes of submission flow condition the forecast of
bid and ask submission fields over the next $\Delta w=300$ seconds.

The Base models train on $10$ trading days from June 8--22, 2026.
The Large models extend this split with $19$ earlier days from
May 11--June 5, giving $29$ training days in total. Training uses window-grid
offsets of $0$, $60$, $120$, $180$, and $240$ seconds, yielding $96{,}300$
Base and $279{,}270$ Large training pairs. Volume-transform and
summary-feature statistics are fitted
separately for each training split using only its unshifted windows.

All models are compared on a fixed corpus from June 23, 2026 containing
$1{,}950$ conditions, with $65$ per stock. Evaluation uses the unshifted grid,
with target windows starting at 10:35 through 15:55 Eastern Time. This day
also serves for checkpoint selection, so the reported results are
validation-set comparisons following chronological training. June 24 is
reserved in the configured test split and is not used in these comparisons.
The same $30$ stocks appear in training and evaluation; per-stock results
characterize performance variation across this observed panel.

\subsection{Experimental Comparisons and Baselines}
\label{sec:exp-comparisons}
The primary comparison uses three Base variants of the conditional diffusion
model in Section~\ref{sec:model-net}. \emph{Base-MSE} trains with the
diffusion MSE in~\eqref{eq:mse}; \emph{Base-UOT} adds the transport
auxiliary objective in~\eqref{eq:total}. Throughout, UOT denotes this
combined MSE+UOT objective. \emph{Base-GRPO} fine-tunes Base-UOT as described
in Section~\ref{sec:model-grpo}. These variants examine how the training
objective and fine-tuning procedure affect field accuracy and predictive
uncertainty, alongside the statistical references below.

An expanded configuration adds \emph{Large-MSE} and \emph{Large-UOT} to
check whether the MSE/UOT findings persist with greater capacity, more data,
and longer training (Section~\ref{sec:exp-models}).

Two parametric baselines fit the observed submission field $A_k^-$.
\emph{Skew-normal} fits one skew-normal distribution per side, using volume
weights within $50$ ticks of that side's modal price. \emph{Skew-normal
mixture} fits the full observed support, with BIC selecting one or two
components per side ($K_{\max}=2$) and observed share volume as the effective
sample size. Both fit separately at each forecast boundary and scale the
fitted shape to the observed side's total volume, yielding a deterministic
forecast field. They use recent flow without book-depth conditioning;
their comparison includes differences in fitting support and component
limits.

For a historical distributional reference, \emph{Empirical} samples
complete bid--ask fields from the same instrument and five-minute window
index on the Base configuration's $10$ training days, using unshifted
windows only. Each selected field is translated from its own snapshot-mid
anchor to the current anchor. Draws are uniform and independent with
replacement. This baseline uses instrument identity, time of day, and
the snapshot anchor; it does not use recent flow or book depth.

\subsection{Training}
\label{sec:exp-models}
\paragraph{Diffusion training.}
Base has $58.7$M parameters (hidden width $d=512$, $12$ blocks, $8$ heads)
and trains for $30{,}000$ steps at learning rate $2\times10^{-4}$; Large
uses $174.3$M parameters ($d=768$, $16$ blocks, $12$ heads) and $100{,}000$
steps at $10^{-4}$. Within each size, MSE and UOT share architecture, data,
and optimizer settings. Both use AdamW with constant learning rates,
batch size $64$, weight decay $0.01$,
gradient-norm clipping at $1$, mixed precision, and EMA decay $0.9999$.
Both sizes use patch width $P=16$, feed-forward width $4d$, and dropout
$0.1$; the grid, diffusion schedule, and UOT coefficients are also shared.
Each learned configuration uses one training seed, $42$.

\paragraph{GRPO fine-tuning.}
Base-GRPO starts from the selected Base-UOT EMA weights and uses the same
training split. Each of $500$ iterations samples $G=8$ fields for each of
$16$ conditions, followed by one policy-update epoch. Updates use AdamW
with learning rate $10^{-5}$, zero weight decay, gradient-norm clipping at
$1$, mixed precision, and EMA decay $0.99$. The policy-ratio clipping
threshold is $0.2$ and the KL-penalty coefficient is $0.02$.

\subsection{Evaluation and Reporting Protocol}
\label{sec:exp-protocol}
All models are evaluated on the common lag0 corpus in
Section~\ref{sec:exp-data}, with matched forecast boundaries and realized
targets. Their conditioning inputs follow Section~\ref{sec:exp-comparisons}.
\paragraph{Forecast generation.}
Diffusion models generate $R=4$ independent fields per condition from
selected EMA weights (evaluation base seed $0$). MSE/UOT use $50$ DDIM
steps ($\eta=0$); Base-GRPO uses its $20$-step training sampler ($\eta=1$).
Empirical also draws four independent fields; each skew-normal baseline
is deterministic ($R=1$). Base/Large share the $50$-step sampler, whereas
the GRPO comparison combines fine-tuning and a sampler change.

\paragraph{Scoring and calibration.}
Fields in shares are scored against full-support targets, including
realized flow outside the model grid, using the eight quantities in
Section~\ref{sec:metric}. Field costs and errors average over defined draws;
distributional scores use the ensemble jointly. Deterministic $R=1$
forecasts have zero inter-draw terms, leaving only target distance.
For GRPO, we retain both energy scores' target and inter-draw components.
Volume calibration compares $R=4$ models using RI and interior-rank
coverage (nominal $0.6$). Ties are broken uniformly at random using
reproducible sample-based seeds.

\paragraph{Aggregation and reporting.}
\label{sec:metric-agg}
Sided $W_1$ and NTVE average defined draws within sides, then equally across sides, requiring both means.
Full-field cost sums sides; log-field MSE follows~\eqref{eq:metric-rho-l2}.
Per-stock and pooled means equally weight defined boundaries, with per-metric valid and excluded counts; volume ranks pool boundary--side pairs.
Raw costs can be dominated by high-volume windows and are interpreted alongside shape and relative- or log-volume metrics.
Comparisons are descriptive, using the checkpoint-selection day and one training seed per learned configuration.
\section{Results}
\label{sec:results}

% Numerical provenance (paths relative to the Finance_World_Model repository).
% Evaluation run:
% evaluation_results/2026-09-22/eval__20260914_dow30_v3_large__cfg6e4678be5d97__20260922T024737Z/
% results.csv SHA256: 333d221b4d9cda7e00d9a7ffd716c8216c8c69d8918c36e7de1c999604303e1d
% metric_results.csv SHA256: c2d687c2a63e0ec88a0e9033a0cc312b69af01b3580d4b873f1bab6e14e6f5bb
% per_sample_metrics.parquet SHA256: d1ca166a364b1f2ad9bddf9ee7107903076a5e9be55fb6cbb59615e26591bb80
% Paper labels -> registered model IDs:
% Skew-normal: skewnorm__band
% Skew-normal mixture: skewnorm_mixture__kmax2
% Empirical: naive_flow__empirical__dow30_v2
% Base-MSE: order_flow_dit__dit__dow30_v2__s42__cfg75175188ab25
% Base-UOT: order_flow_dit__dit__dow30_v2__s42__cfge91d5d1c3334
% Base-GRPO: order_flow_dit_grpo__grpo__dow30_v2__s42__cfg394eed77cff6
% Large-MSE: order_flow_dit__dit__dow30_v3__s42__cfge223a305d4e5
% Large-UOT: order_flow_dit__dit__dow30_v3__s42__cfg39412da9cf87
% All seven field/score means were checked against the per-sample artifact;
% all have 1,950 finite values/model on identical sample IDs (65/stock).
% The GRPO table's D terms are full off-diagonal inter-draw means (not D/2).
% Per-stock counts below compare means over each stock's same 65 conditions.
% Percent changes use unrounded values; displayed values use three decimals.

We assess field accuracy and predictive-distribution quality on the common
$1{,}950$ lag0 conditions across $30$ stocks described in
Section~\ref{sec:exp-data}. All field costs and distributional scores are
defined on every condition for the eight selected models, with no
exclusions. Volume-rank diagnostics use $3{,}900$ condition--side pairs per
model. As specified in Section~\ref{sec:exp-protocol}, these are descriptive
comparisons on the checkpoint-selection day, with one training seed per
learned configuration.

\subsection{Main Comparison with Statistical Baselines}
\label{sec:results-main}

\begin{table*}[t]
    \caption{Main comparison on $1{,}950$ conditions.
    All seven metrics are lower-is-better; bold marks the lowest displayed
    value in each column. The full-field cost is reported in
    $10^6$ share$\cdot$ticks and sided $W_1$ in ticks.
    CRPS denotes log-total-volume CRPS. The mixture uses $K_{\max}=2$.
    The $R=1$ rows are point-mass forecasts.}
    \label{tab:main-results}
    \centering
    \small
    \setlength{\tabcolsep}{3.5pt}
    \begin{tabular}{@{}lrrrrrrrr@{}}
        \toprule
        Model & $R$ & $d_{\mathrm{full}}/10^6$
        & $\mathrm{ES}_T$ & $\mathrm{ES}_2$ & CRPS
        & Sided $W_1$ & NTVE & $\mathrm{MSE}_{\log}$ \\
        \midrule
        Skew-normal & 1 & 450.515 & 14.995 & 0.794 & 0.317
        & 733.904 & \textbf{0.328} & 0.722 \\
        Skew-normal mixture & 1 & 109.510 & 4.154 & 1.452 & 0.317
        & 191.888 & \textbf{0.328} & 2.433 \\
        Empirical & 4 & 95.698 & 2.018 & 0.496 & 0.283
        & 166.599 & 0.756 & 0.934 \\
        \midrule
        Base-MSE & 4 & 110.829 & 2.152 & 0.414 & 0.460
        & 149.220 & 1.058 & 0.656 \\
        Base-UOT & 4 & \textbf{60.074} & 1.591 & \textbf{0.387} & 0.283
        & 123.552 & 0.447 & 0.599 \\
        Base-GRPO & 4 & 62.025 & \textbf{1.494} & 0.409 & \textbf{0.243}
        & \textbf{107.705} & 0.382 & \textbf{0.549} \\
        \bottomrule
    \end{tabular}
\end{table*}

Base-UOT improves both field accuracy and predictive-distribution quality
over Base-MSE (Table~\ref{tab:main-results}). All seven reported metrics
decrease, including a $45.8\%$ reduction in full-field cost and a $26.1\%$
reduction in transport energy score $\mathrm{ES}_T$. The gains also extend
to the joint-field score $\mathrm{ES}_2$, whose binwise geometry differs
from transport, and to volume CRPS.

The statistical baselines place these improvements in context. Base-UOT
outperforms Empirical in full-field cost and both energy scores, and has
the lowest $\mathrm{ES}_2$ among the six models in
Table~\ref{tab:main-results}. Other measures favor different forecasts:
the skew-normal baselines attain the lowest NTVE. The learned forecasts'
advantages in
field accuracy and distributional scores therefore coexist with room for
improvement in marginal volume calibration.

The UOT gains are consistent across the evaluated stock panel. Relative to
Base-MSE, Base-UOT lowers both full-field cost and $\mathrm{ES}_T$ for
every stock. It also lowers full-field cost relative to Empirical on $29$
stocks and $\mathrm{ES}_2$ on all $30$. Each comparison uses the same $65$
conditions per stock, showing that the pooled improvements are accompanied
by gains across the instruments represented in training and evaluation.

\begin{figure}[t]
    \centering
    \includegraphics[width=\columnwidth]{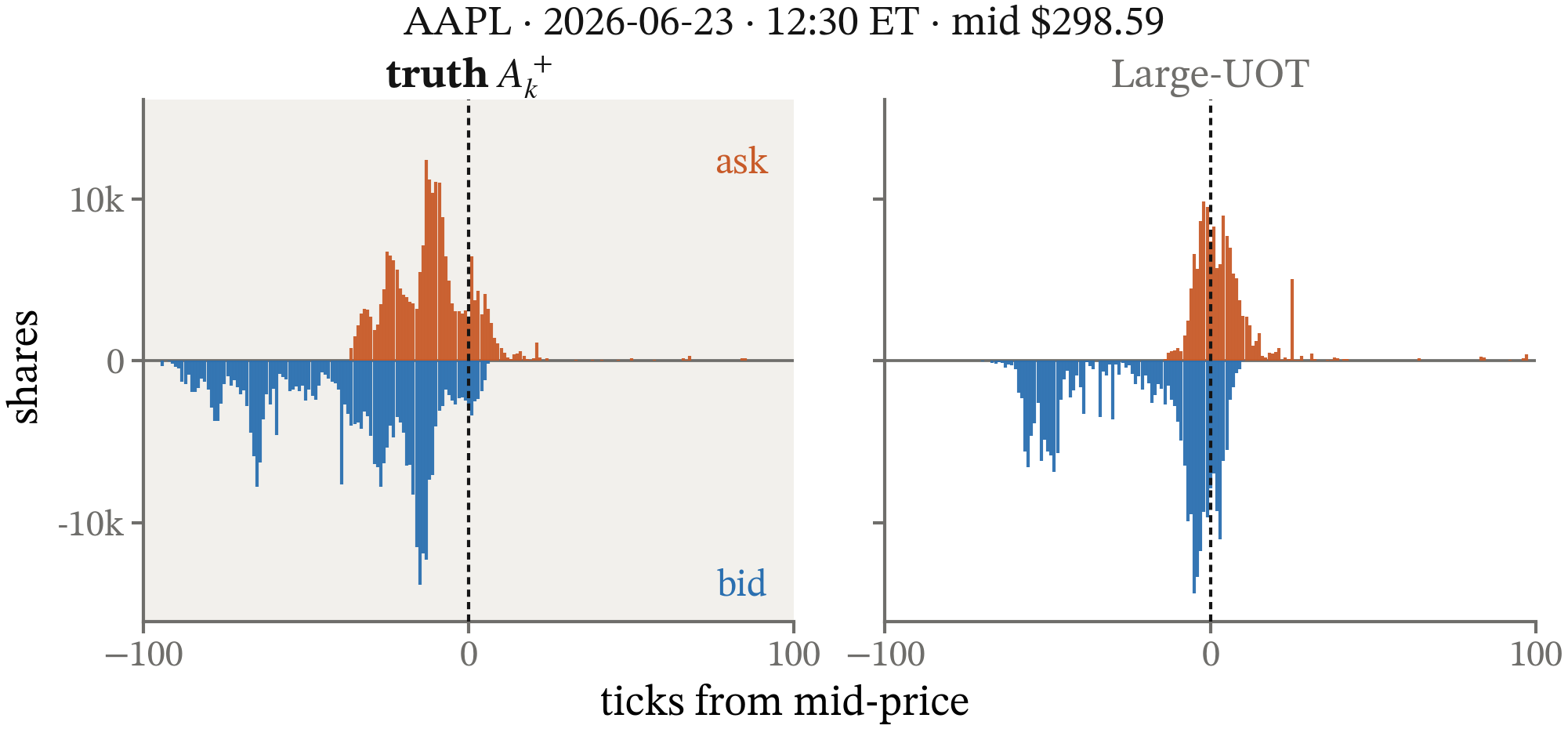}
    \caption{Realized and Large-UOT submission fields within $100$ ticks of
    the mid-price on identical axes: ask shares upward, bid downward.
    The condition is nearest the median Large-UOT energy score among
    mid-session conditions on this instrument. Total volumes are $1.94$M
    and $1.49$M shares, respectively; only roughly a quarter of each field's
    volume lies in the displayed range.}
    \Description{Two panels on identical axes spanning one hundred ticks either
    side of the mid-price. The left panel is the realized submission field, with
    ask volume above a zero line and bid volume below it; the right panel is one
    draw from the Large-UOT model. Both show an ask mode just above the mid-price
    and bid volume spread further below it, of comparable height and width.}
    \label{fig:truth-vs-model}
\end{figure}

Figure~\ref{fig:truth-vs-model} illustrates a generated field by placing
one Large-UOT draw beside the realization at the same condition.

% Condition picked by a pre-stated rule, not by appearance: among mid-session
% conditions on this instrument, the one nearest the median of Large-UOT's own
% energy score. Panels come from the scored prediction bundle (pred_8cd1630bcd8f);
% provenance in studies/20260930_icaif_results_figure/.

\subsection{GRPO: Accuracy, Diversity, and Calibration}
\label{sec:results-grpo}

\begin{table}[t]
    \caption{GRPO comparison on the same lag0 conditions ($R=4$).
    For each energy-score geometry $j\in\{T,2\}$, $T_j$ is the mean
    target distance and $D_j$ the mean off-diagonal inter-draw distance,
    so $\mathrm{ES}_j=T_j-D_j/2$.
    Interior-rank coverage has nominal value $0.600$.
    Distance components and coverage are diagnostics, without a common
    lower-is-better direction.}
    \label{tab:grpo-results}
    \centering
    \small
    \setlength{\tabcolsep}{5pt}
    \begin{tabular}{@{}lrr@{}}
        \toprule
        Quantity & Base-UOT & Base-GRPO \\
        \midrule
        $\mathrm{ES}_T$ & 1.591 & 1.494 \\
        Target distance $T_T$ & 2.897 & 2.510 \\
        Inter-draw distance $D_T$ & 2.611 & 2.033 \\
        \addlinespace
        $\mathrm{ES}_2$ & 0.387 & 0.409 \\
        Target distance $T_2$ & 0.722 & 0.686 \\
        Inter-draw distance $D_2$ & 0.670 & 0.553 \\
        \midrule
        $d_{\mathrm{full}}/10^6$ & 60.074 & 62.025 \\
        Volume CRPS & 0.283 & 0.243 \\
        RI & 0.256 & 0.269 \\
        Interior-rank coverage & 0.395 & 0.385 \\
        \bottomrule
    \end{tabular}
\end{table}

The GRPO variant improves transport and volume scores, but the gain does
not extend to the joint-field score. Relative to Base-UOT, Base-GRPO lowers
$\mathrm{ES}_T$ by $6.1\%$ and volume CRPS by $14.0\%$, while increasing
$\mathrm{ES}_2$ by $5.7\%$ and full-field cost by $3.2\%$.
The pooled full-field cost increases despite decreases in $24$ of the $30$
per-stock means: larger deteriorations elsewhere outweigh those gains.

The energy-score decomposition in Table~\ref{tab:grpo-results} explains
the disagreement between the two scoring geometries. Under both distances,
generated fields become closer to the realization on average, but also
closer to one another. Since $\mathrm{ES}_j=T_j-D_j/2$, the transport
score improves because the reduction in target distance exceeds half the
reduction in inter-draw distance. For the joint-field score, the smaller
target-distance improvement is outweighed by the loss of the dispersion
contribution. A lower average distance to the target therefore need not
yield a better score for the predictive distribution.

Volume-rank diagnostics show a related limitation. RI rises from $0.256$
to $0.269$, while interior-rank coverage falls from $0.395$ to $0.385$,
farther below its nominal value of $0.600$. Thus the improved transport
score and volume CRPS coexist with reduced ensemble dispersion and weaker
pooled volume calibration. These results describe the combined change in
weights and sampling: Base-GRPO uses $20$ DDIM steps with $\eta=1$, whereas
Base-UOT uses $50$ steps with $\eta=0$.

\subsection{Expanded Training Configuration}
\label{sec:results-scaling}
The UOT advantage over MSE in full-field cost and both energy scores
persists in the expanded configuration. Relative to Base-UOT, Large-UOT
reduces full-field cost by $2.1\%$ and $\mathrm{ES}_T$ by $17.5\%$.
Large-MSE nevertheless retains lower NTVE and log-field MSE than Large-UOT.
Capacity, training data, optimizer steps, and learning rate change
together, so this comparison does not isolate their individual effects.

\section{Conclusion}
\label{sec:conclusion}

We presented OrderFlowDiT for conditional generation of unnormalized
limit-order submission fields, together with an evaluation framework that
assesses price placement, total volume, and predictive uncertainty. On
$1{,}950$ validation conditions across $30$ U.S. equities, Base-UOT reduces
full-field transport cost by $45.8\%$ and the transport energy score by
$26.1\%$ relative to Base-MSE, with gains in both measures for every stock.
The UOT advantage persists when model capacity, training data, and
optimization budget increase together. The GRPO variant illustrates the
need for complementary distributional diagnostics: its lower transport
energy score coexists with reduced sample diversity, a worse joint-field
score, and weaker volume calibration. These results support training and
evaluation that account jointly for price geometry, submitted volume, and
the distribution of possible outcomes.

The central limitation is that generating submission flow does not
complete a financial world model. An impact model is also needed to
connect the current book, order flow, and trading actions to the
distribution of subsequent book states and prices. Our aggregated fields
omit cancellations, executions, and event ordering, so they cannot by
themselves determine these transitions. Learning and validating this link,
including the market's response to trading actions, is the next step toward
multi-step simulation and the evaluation of trading decisions.

\bibliographystyle{ACM-Reference-Format}
\bibliography{references}

\end{document}